\documentclass[11pt]{article}

\usepackage[final]{acl}

\usepackage{times}
\usepackage{latexsym}

\usepackage[T1]{fontenc}

\usepackage[utf8]{inputenc}

\usepackage{microtype}

\usepackage{inconsolata}

\usepackage{graphicx}
\usepackage{multirow}
\usepackage{multicol}
\usepackage{booktabs}
\usepackage{array} 
\usepackage{makecell}
\usepackage{threeparttable}
\usepackage{pifont}
\usepackage{amsmath}
\usepackage{colortbl} 
\usepackage{arydshln}
\usepackage{xcolor}

\title{LaSR: Context-Aware Speech Recognition via Latent Reasoning}

\author{%
Heyang Liu$^{1,2}$ \quad Ziyang Cheng$^{1,2}$ \quad Jiayi Huang$^{1}$ \quad Wenyang Xiao$^{1}$  \\ 
\textbf{Ronghua Wu}$^2$ \quad  \textbf{Qunshan Gu}$^{2}$  \quad \textbf{Yanfeng Wang}$^1$ \quad \textbf{Yu Wang}$^{1\dagger}$\\
$^1$Shanghai Jiao Tong University \quad $^2$ Ant Group \\
\texttt{\{liuheyang,muye12,hjy-sjtu,onesheep,wangyanfeng622,yuwangsjtu\}@sjtu.edu.cn}\\
\texttt{\{r.wu, guqunshan.gqs\}@antgroup.com}
}

\begin{document}
\maketitle
\begin{abstract}

Speech recognition in specialized domains requires leveraging contextual or topical information to improve the recognition of domain-specific entities. Speech Large Language Models (Speech LLMs) have substantially advanced speech understanding and reasoning capabilities, making context-aware speech recognition possible without predefined bias lists. In this paper, we propose \textbf{LaSR} (\textbf{La}tent \textbf{S}peech \textbf{R}easoning), a novel training paradigm featuring a context-aware reasoning trajectory that leverages the latent reasoning process. Instead of generating explicit intermediate tokens, LaSR aligns chain-of-thought (CoT) supervision around the acoustic feature region of the target word, and introduces latent reasoning periods for context information grounding and transcriptional transition. Furthermore, to effectively benchmark context-aware speech recognition, we propose Spoken Darwin-Science, a large-scale corpus focusing on academic terminologies. Preliminary experiments on Fun-Audio-Chat demonstrate that LaSR significantly improves terminology recognition without introducing additional latency and consistently outperforms standard supervised fine-tuning baselines. Our findings highlight the potential of latent reasoning in building efficient, context-aware speech assistants.

\end{abstract}

\section{Introduction}

Recent advances in speech large language models (Speech LLMs) have demonstrated impressive capabilities in the understanding and generation of spoken language~\cite{wang2025vocalnet, team2025fun, xu2025qwen3}. These models capture long-range dependencies in audio, leverage contextual cues, and perform reasoning over complex speech inputs. However, the performance of existing models is constrained by intrinsic limitations. On one hand, directly adopting Chain-of-Thought (CoT) techniques often results in limited stability and can even degrade the general intelligence~\cite{li2025reinforcement, xu2025qwen3, wu2025step, comanici2025gemini}. On the other hand, explicit generation of intermediate reasoning steps introduces substantial computational overhead and latency, which is unacceptable in real-time transcriptions and conversations, negatively impacting user experience. Therefore, it is necessary to simultaneously achieve contextual reasoning enhancement of the Speech LLMs without additional computational delay.

\begin{figure}[t]
  \centering
\includegraphics[width=0.47\textwidth]{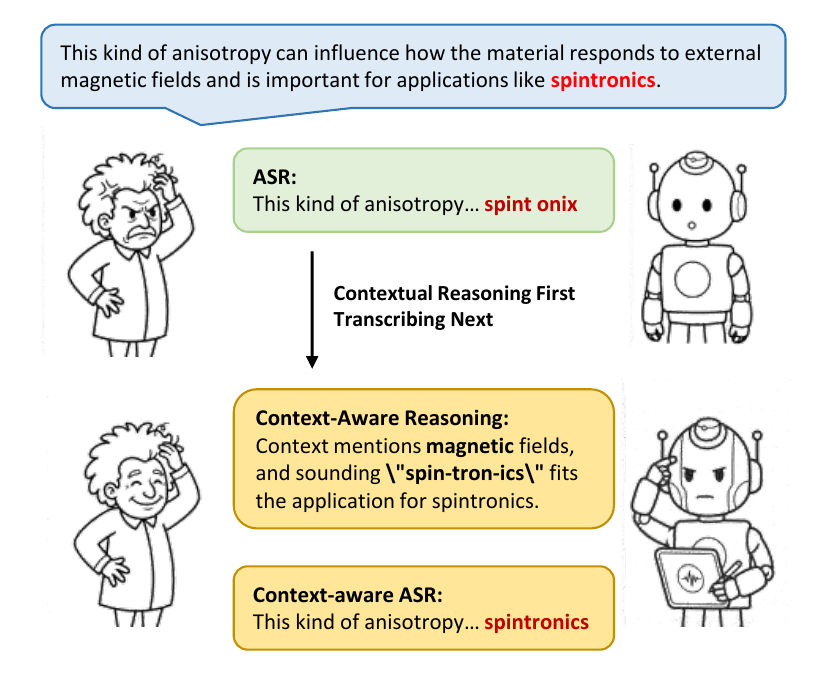}
  \caption{Context-aware ASR perceives the context for reasonable transcription without a predefined bias list.}
  \label{fig:intro}
\end{figure}

Contextual automatic speech recognition (CASR) extends the basic framework by incorporating the bias list, a curated set of task-relevant terms typically ranging from dozens to thousands of entries~\cite{sun2021tree, yang2024ctc, kong2025contextual, deng2026speech}. By explicitly increasing the recognition probability of these target words, CASR achieves improved accuracy for hot words in downstream applications. However, bias lists are not always available in practice, particularly in general-purpose specialized domains. List-free contextual ASR, or context-aware ASR, addresses this limitation by leveraging the contextual reasoning capabilities of large language models (LLMs). As illustrated in Figure~\ref{fig:intro}, this approach performs contextual reasoning to identify semantically relevant target words without relying on an explicit keyword list. In this paper, we construct Spoken Darwin-Science, a large-scale synthesized corpus consisting of broad terminologies, and corresponding real-world speech evaluation sets from online resources. From mainstream ASR models and Speech LLMs, we demonstrate that supervised fine-tuning (SFT) with high-quality synthetic data can effectively improve the model's generalization on real recordings. Building on this, we propose \textbf{LaSR} (\textbf{La}tent \textbf{S}peech \textbf{R}easoning), a training scheme applicable to speech models that utilizes CoT to enhance contextual reasoning capabilities without additional latency and improve the recognition of hard terminologies. Specifically, LaSR follows a listening-while-thinking manner, proposes to align the critical contextual reasoning procedure with the acoustic feature inputs of targeted terminologies, and performs latent reasoning before and after the supervision of the context-aware CoT stage. It addresses the latency overload of explicit CoT outputs before transcriptions and exceeds standard SFT baselines. Our contributions are as follows:



\begin{itemize}
    \item We develop a challenging suite for contextual reasoning capabilities of Speech LLMs. The model is developed to enhance the recognition of academic terminologies through the analysis of previous context.
    
    
    \item We propose LaSR, a training strategy that injects intermediate reasoning supervision and latent reasoning periods into audio encoding periods, allowing Speech LLMs to capture contextual dependencies and reasoning cues without explicit CoT outputs.  
    
    \item Extensive experiments have demonstrated that LaSR successfully leverages the reasoning trajectory and inherent latent reasoning, and validated its effectiveness and efficiency. 
    
\end{itemize}
\section{Related Works}



\subsection{CoT Prompting and Latent Reasoning}

The enhancement of reasoning capabilities is a critical feature of LLM advancement. Early approaches improve model reasoning by explicitly providing intermediate reasoning steps, represented by CoT prompting~\cite{wei2022chain}. By exposing to structured solution trajectories, these methods have been shown to improve performance on arithmetic, logical, and commonsense reasoning benchmarks~\cite{yue2024mammoth, ye2025limo}. Recently, latent reasoning has received increasing attention without explicit intermediate outputs. COCONUT~\cite{hao2024training} breaks down the reasoning procedure into fixed steps. It gradually removes the explicit CoT steps at the front positions, replacing them with autoregressive propagations of the hidden states, thereby providing a wider beam size decoding space. CODI~\cite{shen2025codi} incorporates self-distillation, using the model with explicit CoT as the teacher and latent reasoning as the student, aligning their distribution of generated responses. The compression of explicit CoT is another approach. CoLaR~\cite{tan2026think} samples the compressed CoT tokens at different ratios, while Token Assorted~\cite{su2025token} uses an external VQ-VAE to obtain latent CoT tokens for direct model training. While achievements in text modalities, their effect on speech has not been verified.


\subsection{Contextual ASR}

CASR requires the model to perceive the speaker's context and perform favored transcriptions of a bias list. Traditional ASR methods mainly employ two approaches: biased word probability enhancement based on Weighted Finite-State Transducer (WFST)~\cite{zhao2019shallow}, and explicitly adding a bias encoder to inject relevant vocabulary~\cite{pundak2018deep}. The initial LLM-based attempt employed an additional hot words prompt, explicitly injecting a fixed-size bias list into the LLM backbone~\cite{yang2024ctc}. These approaches all require biased lists (hot words), whether based on training word frequency or simulated user definitions, ignoring the ability of LLMs to perceive and reason about contexts. In \citet{deng2026speech}, CoT-ASR is proposed to produce higher-quality transcriptions by generating analytical reasoning first. This fully utilizes the generative reasoning capabilities of LLMs without predefined bias lists, achieving context-aware speech recognition, while the additional CoT output results in excessive decoding latency and cannot support real-time transcriptions. In contrast, LaSR proposes to influence the latent thinking process streamingly, thereby improving the model's contextual perception and rare-word transcriptions without additional latency.
\begin{figure*}[ht!]
  \centering
\includegraphics[width=0.88\textwidth]{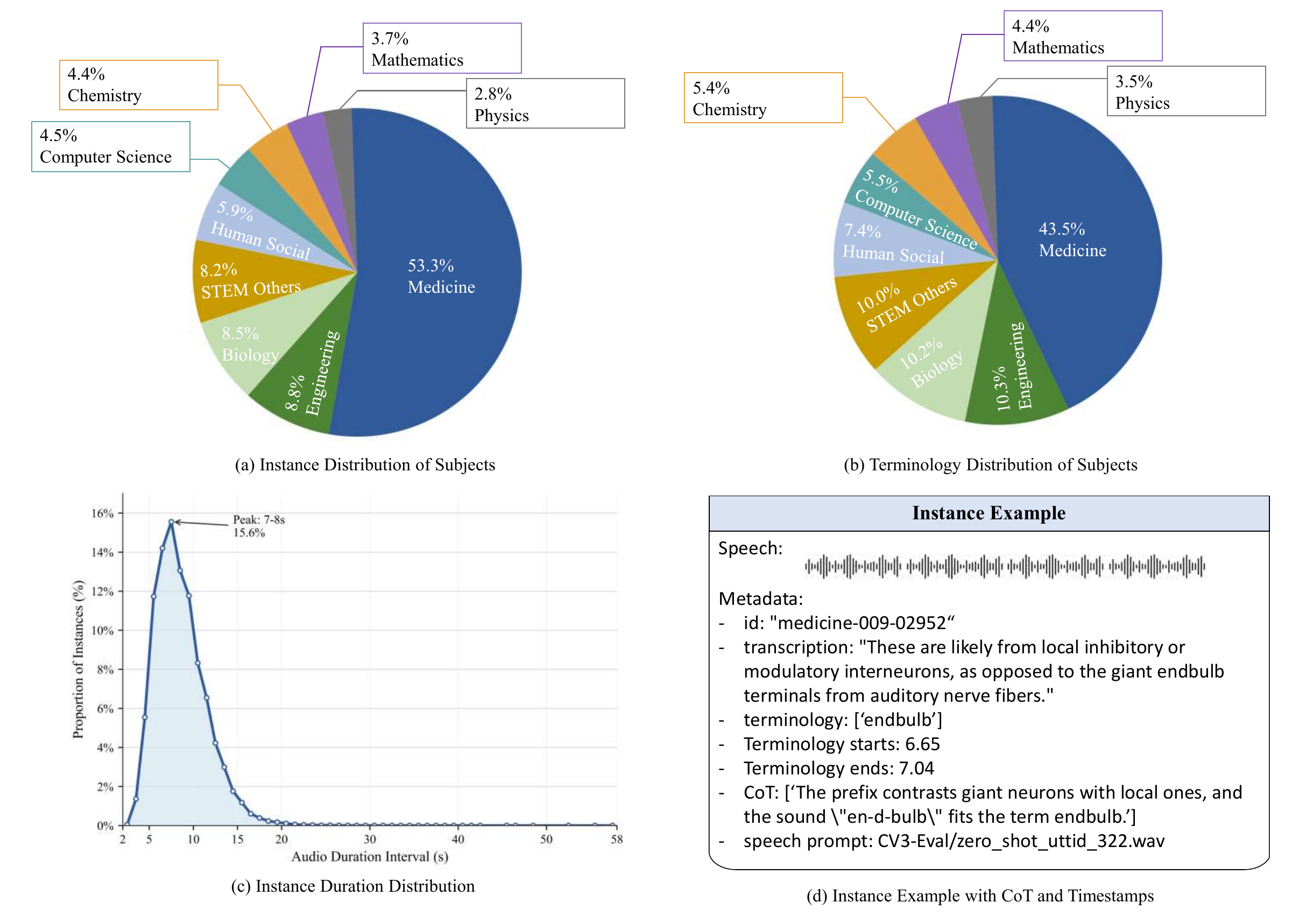}
  \caption{Dataset statistics of Spoken Darwin-Science 20\% subset. (a): The instance distribution of various subjects; (b) The distinct terminology distribution of various subjects; (c) The duration distribution of instances; (d) Instance example with timestamps and CoT annotations.}
  \label{fig:dataset}
\end{figure*}

\section{Spoken Darwin-Science}

In this section, we introduce the corpus proposed in the experiment, including the design principles, main features, construction process, and statistics.

\subsection{Data Principles}

Previous work on CASR typically relied on general conversational corpora (e.g., LibriSpeech, GigaSpeech~\cite{panayotov2015librispeech, chen2021gigaspeech}), and constructed test sets by selecting low-frequency words from these corpora to evaluate context-dependent recognition~\cite{sun2021tree, cui2025exploring}. However, the scaling of recent speech models, particularly Speech LLMs, has diminished the effectiveness: abundant high-quality conversational data in training reduces the scarcity of rare words, making these test sets less challenging. To address this limitation, we focus on the scientific domain comprising academic terminology. These scientifically named entities are highly specialized and often require models to leverage both local acoustic cues and broader contextual information for accurate transcription. Consequently, this dataset directly reflects the contextual reasoning capabilities of speech LLMs.

\begin{table}[t]
  \centering
  \resizebox{\linewidth}{!}{%
  \begin{tabular}{l l} 
    \hline
    \textbf{Subject} & \textbf{Terminology} \\
    \hline
    Computer Science & biqubits, chainwork \\
    Engineering      & suffusion, rhizotron \\
    Human Social     & floristry, chieftaincies \\
    Medicine         & flavonolignans, phosphatide \\
    Biology          & abacopterin, dysmorphogenesis \\
    Chemistry        & phenyltrimethoxysilane, pentanethiol \\
    Math             & polynomiography, contactomorphic \\
    Physics          & graviscalars, nonpolarizing \\
    STEM (Others)    & nakhlites, equatorwards \\
    \hline
  \end{tabular}%
  }
  \caption{Example terminology of different subjects of Spoken Darwin-Science.}
  \vspace{-2mm}
  \label{tab:terminology}
\end{table}

\subsection{Data Construction}

\subsubsection{Terminology Definition}

Terminologies represent highly specialized vocabularies that rarely appear in daily conversations but play a crucial role in semantic understanding and academic dialogue. We refer to the vocabulary from the GigaSpeech-XL training set and calculate word frequencies~\cite{chen2021gigaspeech}. GigaSpeech is a massive speech dataset with over 10,000 hours of transcribed text, primarily sourced from YouTube, Podcast, and Audiobook, and most are daily words. We defined words appearing less than 10 times as terminologies,  which is validated through manual verifications. 

\subsubsection{Training Set}

We select academic papers from various disciplines as our initial sources. Specifically, the corpus is sourced from Darwin-Science~\cite{qin2026data}, a high-quality collection of papers that has undergone multiple stages of text cleaning, covering nine scientific fields, including biology, chemistry, and human social. We removed the fixed-length characters at the beginning and end, retaining only the main text, and segmented it into sentences using NLTK~\cite{bird2006nltk}. A single sentence is retained only if it contains terminology, and a maximum of five sentences with the same terminology are kept. The final corpus comprises 2.7M instances filtered from text related to 440B tokens. For speech synthesis, we use Qwen3-TTS-1.7B~\cite{hu2026qwen3}, and clips with DNSMOS Pro~\cite{cumlin2024dnsmos} above 3 from CV3-Eval~\cite{du2025cosyvoice} and seed-tts-eval~\cite{anastassiou2024seed} are selected as the speech prompts to ensure the quality of the synthesized speech. In our experiments, Qwen3-TTS demonstrates strong generalization capabilities, effectively converting words into speech based on their orthographic composition and general pronunciation rules. This is particularly advantageous for synthesizing scientific terminology, which often consists of multiple common phonetic units and frequently forms compound words. Such a systematic structure enables accurate and natural pronunciation even for previously unseen terminology. Terminology examples in our constructed corpus are shown in Table~\ref{tab:terminology}, and more detailed pipeline is summarized in Appendix~\ref{app: dataset_pipeline}.

\subsubsection{Evaluation Set} \label{evaluation_set}

We construct the evaluation set of real-world scientific audio scenarios from publicly available resources. Candidate videos are sourced from YouTube, covering biology, medicine, chemistry, physics, geography, and general sciences. We preserve videos of lectures, popular science courses, documentaries, and TED talks that provide human-annotated subtitles, ensuring high reliability in speech transcription. The video metadata is parsed into timestamped segments and then merged based on intervals and sentence-ending punctuation to form sentence-level audio segments. Those containing terminologies are retained, and we limit the speech length to within 30 seconds. Following this procedure, we construct a total of 2,000 real-world scientific speech segments for evaluation.

\subsubsection{CoT}

Considering the scale of the training corpus, we randomly select 20\% instances (539K) and construct the CoT. To avoid the additional time overhead of explicit reasoning based on full text, we followed a "listen-while-think" strategy, providing the terminology and its preceding text when generating the inference trajectory. Qwen3.5-27B~\cite{qwen35blog} is instructed to infer the context and intent, and then decompose the terminology pronunciation, and finally transcribe the word. The prompt used is shown in Appendix~\ref{app: dataset_pipeline}. In addition, we leverage Qwen3-ForcedAligner~\cite{shi2026qwen3} to generate word-level timestamps in order to determine the time anchor point for model reasoning.

\begin{figure*}[h]
  \centering
\includegraphics[width=1\textwidth]{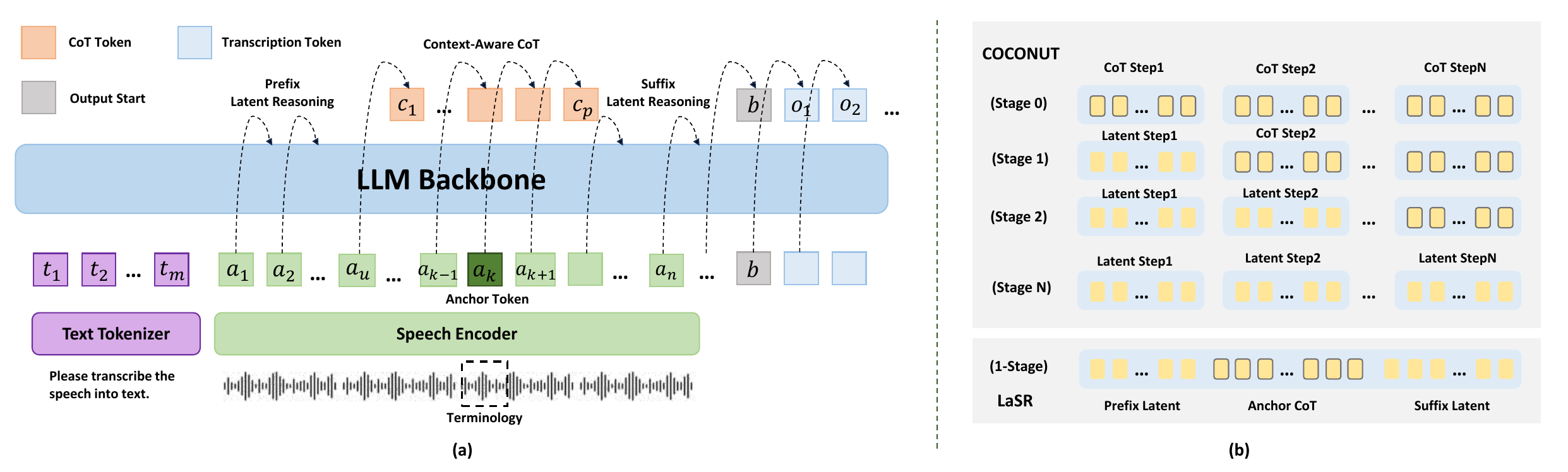}
  \caption{LaSR training method. (a) Structured causal reasoning trajectory of LaSR; (b) Comparison with textual implicit reasoning (e.g., COCONUT).}
  \label{fig:methods}
\end{figure*}

\subsection{Human Verification}

To validate the quality of the terminology speech synthesis, we randomly select 1000 training samples for four humans fluent in English with mainstream online dictionaries as references. Regarding pronunciation accuracy, the human experts consider that 96.8\% of the terminologies are synthesized correctly and with normal pronunciation, on average. This confirms the high quality of Spoken Darwin-Science training corpus. The details of the annotation process are located in Appendix~\ref{app: human_verification}.

\subsection{Dataset Statistics}

The dataset statistics of 20\% Spoken Darwin-Science with CoT annotations are shown in Figure~\ref{fig:dataset}. Regarding the subject and speech duration, the full corpus shares similar distributions. Spoken Darwin-Science consists of 9 main scientific domains. Medicine comprised the majority of the dataset, accounting for 53.3\% of the entries (and a similar proportion of the duration), and contributing 43.5\% of the terminology. The other three major subjects are engineering, biology, and STEM (Others), which each contributed more than 10\% of the terminology, corresponding to more than 8\% of the duration and entries. The remaining five subjects are human social, computer science, chemistry, mathematics, and physics, which together account for approximately 26\% of the terminology and 21\% of the data volume. This distribution is related to the inherent characteristics of these disciplines; some subjects extensively utilize daily language (e.g., mathematics), resulting in fewer highly specialized terms. A more detailed data distribution is shown in Appendix~\ref{app: dataset}.

\section{LaSR}

\subsection{Overview}


LaSR decomposes the single training step into two phases. The continuous speech signals are converted into audio tokens and progressively input into the LLM backbone. LaSR imposes a structured causal trajectory over the audio-token sequence: unlabeled latent audio states first, then timestamp-aligned explicit CoT supervision around the terminology region, unlabeled latent states after the CoT span, and final transcription supervision at the response positions. Unlike text-based approaches that gradually convert explicit CoT into implicit reasoning, LaSR modulates the model’s internal reasoning both before and after key stages by targeting intermediate thought processes. The model’s intrinsic trajectory does not need to be predefined, but rather to execute the necessary reasoning at intermediate stages. 



\subsection{Methodology}




 As shown in Figure~\ref{fig:methods} (a), the Speech LLM is fine-tuned with a text instruction $T$ and a speech input $A$ of duration $d_a$. For the causal LLM backbone, text instruction is tokenized into discrete tokens $(t_1, t_2, ..., t_m)$, and speech input is encoded as features $(a_1, a_2, ..., a_n)$. Different from traditional autoregressive training, the prefill phase is incorporated into the training process to enhance the model's reasoning and thinking process under partial information. Before the acoustic features $a_u$ are input into the backbone, the model performs prefix latent reasoning, perceiving the context along the internal trajectory. Upon input of $a_u$ into the model, the large model backbone is guided to perform explicit CoT reasoning of length $p$, which sequentially summarizes the preceding context, the pronunciation of terminology, and finally yields the grapheme of the terminology. After the explicit reasoning, the model continues with suffix latent reasoning until the speech information is fully processed, naturally transitioning to the subsequent transcription output. At this point, the model generates a response start signal $b$ and produces the corresponding transcribed text $O=(o1, o2,..., o_q)$.

  Under this premise, the starting position of CoT, $c_1$, needs to be determined, corresponding to the audio token $a_u$. For a speech input of duration $d_a$, we define $\tau$ as the terminology start time obtained through forced alignment. Under non-streaming speech encoder conditions, the anchored index $k$ is calculated as:

  \begin{equation}
  k = \left\lfloor \frac{\tau}{d_a} n \right\rfloor
  \end{equation}

In this setting, the anchor token represents the input audio feature when the terminology word has the most significant impact, affecting the time when CoT participates in model supervision. There are two considerations in the design: a) CoT is added after the anchor token to ensure that the model fully begins to perceive the terminology; b) CoT is added before the anchor to ensure that contextual consideration is given first, especially the topic and intent. In the experiment, we evaluate both methods. The former shifted CoT backward by features related to 0.15s, while the latter shifts it forward to 0.50s, or randomly, but ensures that the hidden state corresponding to the terminology falls within the supervision range of CoT.


  In the implementation, consecutive audio placeholder labels are overwritten with the tokenized latent CoT labels until either the CoT sequence is exhausted or the audio suffix ends. Other prompt and audio input positions remain masked, except for the assistant transcription positions of the ASR target token labels. Therefore, the latent reasoning and CoT phase are not generated as part of the visible output sequence. The language-model loss is the standard next-token cross-entropy over the union of the transcription labels and latent CoT labels:
  \begin{equation}
  \mathcal{L}_{\mathrm{LM}}
  = - \frac{1}{L}
  \sum_{i\in \Omega_{\mathrm{ASR}}\cup\Omega_{\mathrm{CoT}}}
  \log p_\theta(y_i \mid x_{<i})
  \end{equation}

 \noindent where $L=|\Omega_{\mathrm{ASR}}\cup\Omega_{\mathrm{CoT}}|$, and $x_i \in \{A, b, O\}$.





\begin{table*}[t]
  \centering
  \resizebox{1\textwidth}{!}{
  \begin{tabular}{lc>{\arraybackslash}c>{\centering\arraybackslash}p{1.7cm}>{\centering\arraybackslash}p{2.5cm}>{\centering\arraybackslash}p{2.7cm}>{\centering\arraybackslash}p{2.5cm}}
  \toprule
    \toprule
{\textbf{Model}} & {\textbf{CoT}} & {\textbf{Anchor}} & {\textbf{WER (\%)}} & {\textbf{EER (Base, \%)}} & {\textbf{EER (Hard, \%)}} & {\textbf{EER (Avg, \%)}} \\ 
\hline
Qwen3-ASR-0.6B & \ding{56} & - & 6.27 & 18.50 & 42.93 & 27.84 \\
\arrayrulecolor[gray]{0.5}
\cdashline{1-7}
\arrayrulecolor{black} 
Qwen3-ASR-1.7B & \ding{56} & - & 5.60 & 10.55 & 30.93 & 18.37 \\
\arrayrulecolor[gray]{0.5}
\cdashline{1-7}
\arrayrulecolor{black} 
Whisper-large-v3 & \ding{56} & - & \textbf{4.50} & \textbf{8.35} & \textbf{30.56} & \textbf{16.86} \\
\hline
 Fun-Audio-Chat-8B & \ding{56} & - & 6.25  & 12.68   & 32.74  & 20.37 \\
\arrayrulecolor[gray]{0.5}
\cdashline{1-7}
\arrayrulecolor{black} 
\multirow{2}{*}{+ SFT}& \ding{56} & - & $6.90_{\textcolor{red}{\hspace{1mm}\scriptsize 0.65\uparrow}}$ & $13.23_{\textcolor{red}{\hspace{1mm}\scriptsize 0.55\uparrow}}$ & $26.64_{\textcolor{green}{\hspace{1mm}\scriptsize 6.10\downarrow}}$ & $18.32_{\textcolor{green}{\hspace{1mm}\scriptsize 2.05\downarrow}}$ \\
& \ding{52} & - & $7.93_{\textcolor{red}{\hspace{1mm}\scriptsize 1.68\uparrow}}$ & $18.83_{\textcolor{red}{\hspace{1mm}\scriptsize 6.15\uparrow}}$ & $28.08_{\textcolor{green}{\hspace{1mm}\scriptsize 4.66\downarrow}}$ & $22.39_{\textcolor{red}{\hspace{1mm}\scriptsize 2.02\uparrow}}$ \\
\arrayrulecolor[gray]{0.5}
\cdashline{1-7}
\arrayrulecolor{black} 
+ CoT-ASR & \ding{52} & - & $6.52_{\textcolor{red}{\hspace{1mm}\scriptsize 0.27\uparrow}}$ & $11.42_{\textcolor{green}{\hspace{1mm}\scriptsize 1.26\downarrow}}$ & $25.13_{\textcolor{green}{\hspace{1mm}\scriptsize 7.61\downarrow}}$ & $16.67_{\textcolor{green}{\hspace{1mm}\scriptsize 3.70\downarrow}}$ \\
\arrayrulecolor[gray]{0.5}
\cdashline{1-7}
\arrayrulecolor{black} 
\multirow{3}{*}{+ LaSR} & \ding{52} & \cellcolor[rgb]{ .902,  .902,  .902}{+ 0.15s} & \cellcolor[rgb]{ .902,  .902,  .902} $6.97_{\textcolor{red}{\hspace{1mm}\scriptsize 0.72\uparrow}}$ & \cellcolor[rgb]{ .902,  .902,  .902} $12.05_{\textcolor{green}{\hspace{1mm}\scriptsize 0.63\downarrow}}$ & \cellcolor[rgb]{ .902,  .902,  .902} $26.64_{\textcolor{green}{\hspace{1mm}\scriptsize 6.10\downarrow}}$ & \cellcolor[rgb]{ .902,  .902,  .902} $17.64_{\textcolor{green}{\hspace{1mm}\scriptsize 2.73\downarrow}}$ \\
& \ding{52} & \cellcolor[rgb]{ .902,  .902,  .902}{- 0.50s} & \cellcolor[rgb]{ .902,  .902,  .902} $\textbf{6.09}_{\textcolor{green}{\hspace{1mm}\scriptsize 0.16\downarrow}}$ & \cellcolor[rgb]{ .902,  .902,  .902} $\textbf{10.79}_{\textcolor{green}{\hspace{1mm}\scriptsize 1.89\downarrow}}$ & \cellcolor[rgb]{ .902,  .902,  .902} $\textbf{25.00}_{\textcolor{green}{\hspace{1mm}\scriptsize 7.74\downarrow}}$ & \cellcolor[rgb]{ .902,  .902,  .902} $\textbf{16.23}_{\textcolor{green}{\hspace{1mm}\scriptsize 4.14\downarrow}}$ \\
& \ding{52} & \cellcolor[rgb]{ .902,  .902,  .902}{- Random} & \cellcolor[rgb]{ .902,  .902,  .902} $6.17_{\textcolor{green}{\hspace{1mm}\scriptsize 0.08\downarrow}}$ & \cellcolor[rgb]{ .902,  .902,  .902} $11.10_{\textcolor{green}{\hspace{1mm}\scriptsize 1.58\downarrow}}$ & \cellcolor[rgb]{ .902,  .902,  .902} $25.13_{\textcolor{green}{\hspace{1mm}\scriptsize 7.61\downarrow}}$ & \cellcolor[rgb]{ .902,  .902,  .902} $16.42_{\textcolor{green}{\hspace{1mm}\scriptsize 3.95\downarrow}}$ \\
\hline


\hline
  \end{tabular}
  }
  \caption{Context-aware ASR performance. The best individual results for each type are highlighted in \textbf{bold}, and absolute performance changes compared to the Base model are denoted by $\textcolor{green}{\hspace{1mm}\scriptsize\downarrow}$ (improved) and $\textcolor{red}{\hspace{1mm}\scriptsize\uparrow}$ (degraded). }
  \label{tab: result1_main}
  \vspace{-2mm}
\end{table*}

\subsection{Latent Reasoning Comparison}

In Figure~\ref{fig:methods} (b), we compare LaSR with textual implicit reasoning, represented by COCONUT. Most prior latent reasoning approaches focus on progressively transforming externally visible reasoning trajectories into internal thought processes. While these methods have demonstrated the potential of latent supervision, they generally rely heavily on a structured reasoning sequence, limiting the model’s ability to explore alternative or emergent reasoning paths. LaSR addresses these limitations by establishing fixed anchor points within the reasoning process, allowing the model to perform inner thinking both before and after these key steps. Specifically, prefix latent reasoning is aligned toward the CoT anchor, guiding the model to capture intermediate reasoning, whereas suffix latent reasoning is aligned toward the final transcription, ensuring that latent reasoning ultimately supports accurate output. Furthermore, empirical studies, particularly in speech tasks, indicate that automatically generated or manually defined CoT sequences can be detrimental to performance~\cite{li2025reinforcement, xu2025qwen3}, as they may misalign with the natural temporal dynamics of audio or constrain the model’s internal representations. LaSR explicitly considers the temporal structure of speech. Without such time-aligned latent supervision, the model may fail to attend to critical acoustic information, rendering intermediate reasoning less effective than directly producing the final transcription. By combining result supervision with time-aligned latent reasoning, LaSR enables a flexible yet structured approach to enhancing context-awareness and reasoning.

\section{Experiment}

\subsection{Experiment Settings}

\subsubsection{Baseline Models}

We first demonstrate that the synthesized Spoken Darwin-Science enables the improvement of terminology recognition of real recordings. We perform SFT with LLM-based ASR models, including Qwen3-ASR-0.6B and Qwen3-ASR-1.7B~\cite{shi2026qwen3}. For context-aware speech recognition, we proceed with Fun-Audio-Chat-8B~\cite{team2025fun} to ensure strong contextual reasoning capabilities. Fun-Audio-Chat leverages existing pre-trained models and undergoes multi-stage post-training, enabling text and speech interactions with speech recognition capabilities. The ASR models are trained using 1 NVIDIA A100 GPU, guided by the official implementation, while Fun-Audio-Chat uses 8 NVIDIA A100 GPUs based on LLaMA-Factory~\cite{zheng2024llamafactory}. In addition, we report on the performance of Whisper-large-v3~\cite{radford2023robust} on the evaluation set, as a reference for absolute performance.


\subsubsection{Training Corpus and Evaluated Metrics}

 For model training, all experiments use a 20\% training subset, and Qwen3-ASR is further trained on the full corpus to validate data quality. For evaluation, as explained in Section~\ref{evaluation_set}, we use real-world recordings, each containing terminology that is difficult for speech models. Based on GigaSpeech-XL word frequencies, we decompose a Base set of 1-9 frequency terminologies and a Hard set containing out-of-vocabulary (OOV) words.  We report the word error rate (WER) for overall transcription accuracy, and the entity error rate (EER) for terminologies~\cite{deng2026speech}. 


\subsection{Supervised Fine-tuning}

The supervised fine-tuning results on Qwen3-ASR are shown in Appendix~\ref{app:sft_exp}. With 20\% of the training corpus, both models show steady improvement in overall transcription and terminology recognition. The WER of the 0.6B model decreased from 6.27\% to 5.83\%, while that of the 1.7B model dropped from 5.60\% to 5.34\%. When increasing to full volume, the EER steadily drops while the WER fluctuates upward, possibly due to a discrepancy between the fine-tuned vocabulary and the general words. This experiment demonstrates that training with Spoken Darwine-Science enables generalization to real-world scientific speech scenarios.




\subsection{LaSR Results}

We evaluate various training strategies, as shown in Table~\ref{tab: result1_main}. Standard SFT improves the hard set and overall terminology, though the base set slightly underperforms. For explicit thinking sequences before transcriptions, we evaluate partial-context CoT identical to LaSR (SFT with CoT in the main table), and full-context CoT generated according to CoT-ASR. For the former, the performance degrades. Specifically, the recognition accuracy of the base set terminology drops sharply, with EER increasing from 12.68\% to 18.83\%. The hard set performance has improved, but still underperforms non-thinking SFT. In contrast, the incorporation of full-context CoT yields consistent performance improvements, with all metrics outperforming standard SFT; notably, the overall EER drops to 16.67\%. This demonstrates that the CoT content is particularly critical for explicit thinking approaches.

For LaSR, we have evaluated 3 distinct CoT timestamps. By delaying the terminology CoT timing by 0.15s, the model backbone aggregates more auditory information. The speech recognition results are similar to simple SFT without thinking, but it achieves better results in base terminology, and the overall terminology EER decreases by 2.73\%. In the second method, the CoT thinking strategy is added 0.50s before the terminology appears, ensuring the backbone perceives intent and speech topic first, then addressing the terminology pronunciation. This configuration achieves optimal performance, with all metrics surpassing CoT-ASR, and the hard-set EER decreases by 7.74\%. Compared with the base model, the proposed method reduced the overall EER by 4.14\%, corresponding to a relative improvement of 20.32\%. This result surpasses the evaluated ASR models in terminology recognition, highlighting the potential in real-world scientific dialogue and interaction scenarios. For the last setting, the CoT start point is randomly placed ahead of the anchor time, while we guarantee that the terminology anchor token remains in the CoT region. All metrics have improved, with the overall WER remaining stable. These experiments demonstrate that LaSR can effectively improve context-aware speech recognition.

\subsection{Context-driven Inference}

\begin{figure*}[h]
  \centering
\includegraphics[width=0.7\textwidth]{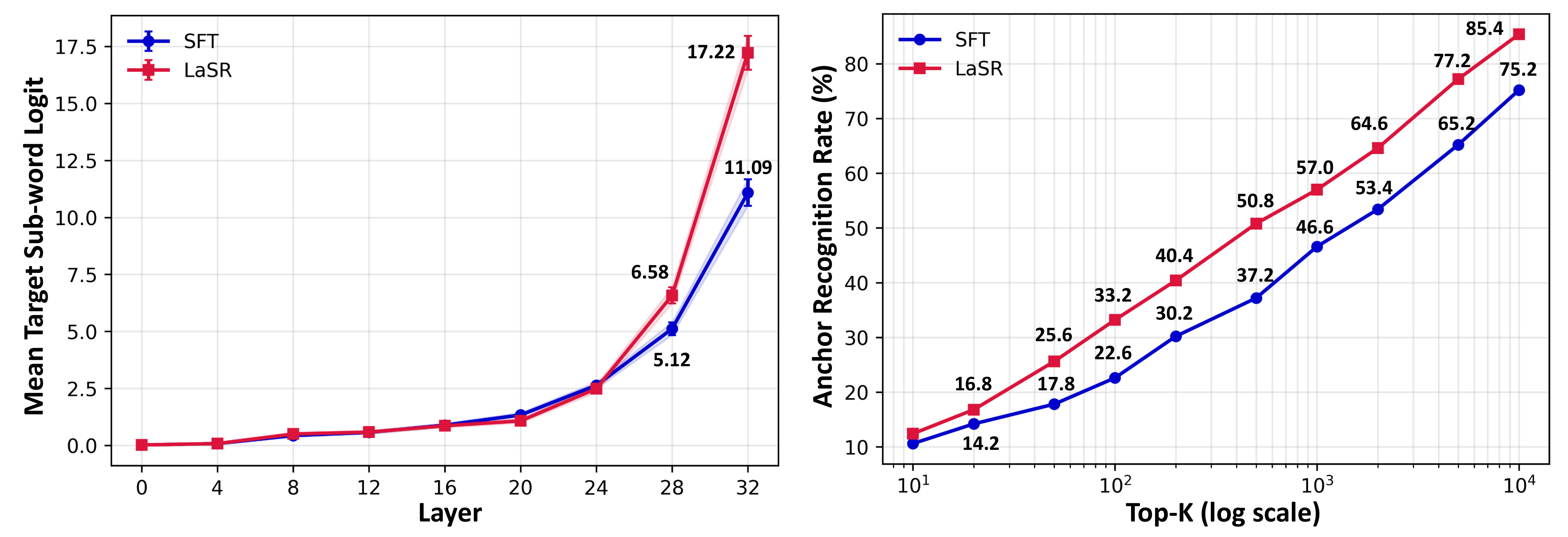}
  \vspace{-3mm}
  \caption{Mechanism analysis. Left: mean target token logits across layers; Right: proportions of target tokens entering Top-k predictions at Layer 32.}
  \label{fig:mechanism}
  \vspace{-4mm}
\end{figure*}

To thoroughly elucidate the underlying mechanism of LaSR, we conduct a rigorous Logit-Lens probing experiment. We randomly sample an evaluation subset of 500 academic terms and generate temporal anchors. We extract the hidden states at the targeted anchor frames from both the standard SFT baseline and the optimal LaSR model. These states are hierarchically projected through the frozen textual LM head to assess the logit values and rank shifts of the target subwords. A non-parametric paired sign test is applied to ensure statistical rigor. As shown in Figure~\ref{fig:mechanism}, in the deeper layers (Layer 28-32), LaSR substantially increases the logit of the target subword, indicating that it preconditions the hidden states toward the targeted semantic token, and the network has already formulated a sharp internal prior. Moreover, at Layer 32, the proportion of samples in which the target terminology subword enters the decoder's Top-$k$ predictions (anchor recognition rate) consistently exceeds that of the SFT baseline across all $k$ settings, with the Top-100 lead exceeding 10.6\%. These results provide direct mechanistic evidence that LaSR fundamentally reshapes the internal representations in the deep layers. More detailed metrics and analyses are provided in Appendix~\ref{app:mechanism}.


\subsection{Zero Shot Performance} \label{subsec:zero_shot_exp}

LaSR eliminates the need to construct corresponding bias lists based on specific scenes and contexts. In essence, the training aims to develop the ability to reason based on context, which allows for direct deployment across various proprietary scenarios. We crawl additional instances for finance and gaming topics, each around 1k sentences, as shown in Table~\ref{tab:zero_shot}. The experimental results demonstrate that despite the absence of specialized training, LaSR still achieves performance improvements compared to standard SFT, especially for hard terms.

\begin{table}[h]
\centering
\resizebox{0.5\textwidth}{!}{
\begin{tabular}{lccccc}
\toprule
\textbf{Domain} & \textbf{Method} & \textbf{WER(\%)} & \textbf{EER(Base,\%)} & \textbf{EER(Hard,\%)} & \textbf{EER(Avg,\%)} \\
\midrule
\multirow{4}{*}{Finance} & Base & \textbf{12.37} & 30.65 & 56.74 & 39.12 \\
 & SFT & 14.16 & 28.19 & 54.42 & 36.71 \\
 & CoT-ASR & 15.52 & 29.98 & 55.35 & 38.22 \\
 & LaSR & 14.15 & \textbf{27.07} & \textbf{53.49} & \textbf{35.65} \\ \midrule
\multirow{4}{*}{Gaming} & Base & 11.95 & 38.27 & 60.36 & 48.44 \\
 & SFT & \textbf{11.70} & 36.35 & 53.60 & 44.29 \\
 & CoT-ASR & 12.81 & \textbf{35.19} & 54.28 & \textbf{43.98} \\
 & LaSR & 12.02 & 37.31 & \textbf{52.03} & 44.09 \\
\bottomrule
\end{tabular}
}
\caption{Performance in Finance and Gaming scenarios.}
\label{tab:zero_shot}
\vspace{-4mm}
\end{table}

\subsection{With Contextual Biasing}




We have evaluated performance with bias lists. In addition to the ground-truth terminology, distractors are randomly selected from the training set terminology vocabulary. The experimental results are listed in Table~\ref{tab:contextual_biasing}. For instances where standard ASR transcriptions cannot be generated due to instruction-following failures, we fall back on a prompt without contextual biasing. Taking the 1000-word bias list as an example, LaSR successfully reduces the failure rate from 1.5\% to 0\%. This demonstrates that the enhancement in reasoning capability directly translates to improved instruction-following performance. In terms of overall performance, contextual biasing exhibits a slightly higher efficacy than LaSR on the 1000-entry bias list due to additional word constraints. Nevertheless, when combined, they achieve a steadier enhancement, culminating in an EER of 13.75\%.

\begin{table}[h]
\centering
\resizebox{0.5\textwidth}{!}{
\begin{tabular}{lccccc}
\toprule
\textbf{Method} & \textbf{Bias List} & \textbf{WER(\%)} & \textbf{Base(\%)} & \textbf{Hard(\%)} & \textbf{Avg(\%)} \\
\midrule
Base & - & 6.25 & 12.68 & 32.74 & 20.37 \\
+ LaSR & - & 6.09 & 10.79 & 25.00 & 16.23 \\
+ CB & 2000 & 9.33 & 10.00 & 26.26 & 16.23 \\
+ LaSR, CB & 2000 & 6.28 & 8.90 & 23.99 & 14.67\\
+ CB & 1000 & \textbf{5.81} & 9.37 & 25.00 & 15.35 \\
+ LaSR, CB & 1000 & 6.21 & \textbf{8.74} & \textbf{21.84} & \textbf{13.75} \\
\bottomrule
\end{tabular}
}
\caption{Performance with contextual biasing (CB).}
\label{tab:contextual_biasing}
\vspace{-4mm}
\end{table}

\subsection{Latency-free Inference}

The thinking-while-listening paradigm of LaSR constrains the CoT period to be synchronous with the speech encoding process, introducing a fundamental trade-off between interaction latency and flexibility. CoT-ASR introduces explicit reasoning after the multimodal input is completely processed. While this allows for flexible, unconstrained CoT exposure to full context, it inevitably introduces significant decoding delays. On the contrary, LaSR achieves zero additional latency and overhead-free absolute improvement. With a single NVIDIA L20 GPU, CoT-ASR introduces an average delay of 1.304 seconds, while LaSR completely solves this problem. Detailed analysis is in Appendix~\ref{app: latency}.



\section{Conclusion}

We explore the bias-list-free CASR of Speech LLMs and propose LaSR, a latency-free latent reasoning strategy. By structuring a causal trajectory for reasoning anchored to the timestamp of the targeted word, LaSR successfully internalizes the context-aware reasoning process. It forces the model to capture intermediate contextual dependencies before and after key acoustic stages without the latency burden of explicit reasoning outputs. To support this research, we construct Spoken Darwin-Science, a comprehensive terminology-centric speech dataset tailored for rigorous contextual evaluation. Our experiments indicate that LaSR yields substantial improvements in recognizing challenging scientific terminologies over supervised fine-tuning and explicit CoT approaches.

\clearpage
\section*{Limitations}

LaSR represents a preliminary exploration of latent reasoning within Speech LLMs. Current experiments are based on the Fun-Audio-Chat framework; although the underlying model architecture is broadly generalizable, it has not yet undergone more extensive empirical validation on other models. Furthermore, the speech encoding process is non-streaming, which implies our anchor token timestamps serve merely as approximations. Finally, LaSR is currently confined to contextual ASR tasks. Extending to speech interaction and audio reasoning would broaden its scope, which would also impose more rigorous demands on the reasoning chain. Finally, the CoT construction scheme, especially the length, affects the period ratio, thus influencing the improvement effect.

\section*{Ethical Considerations}

All the models in our paper are downloaded from publicly released model cards, and we strictly follow the user license. The sourced data is collected from publicly available resources, and we perform speech synthesis using publicly available speech prompts and TTS models. Human verifications are conducted by college students in the author lists. The evaluation set is sourced from online resources and is for academic usage only. The real evaluation data will not be made public, but the corresponding acquisition methods and metadata will be available.

\bibliography{main}

\appendix

\clearpage

\section{Detailed Construction Pipeline} \label{app: dataset_pipeline}

\begin{figure}[h]
  \centering
\includegraphics[width=0.47\textwidth]{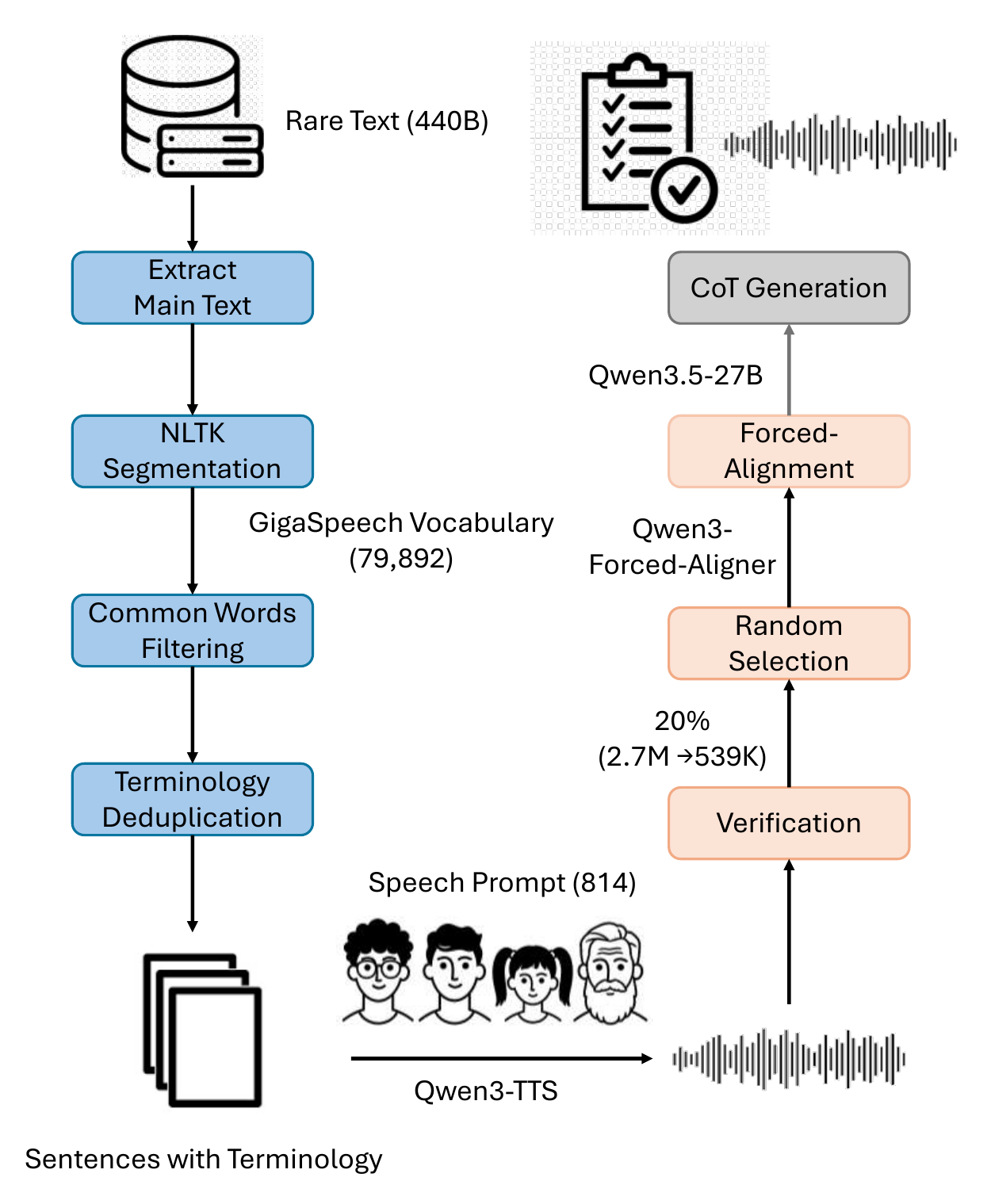}
  \caption{Construction pipeline of Spoken Darwin-Science training set.}
  \label{fig:detailed_pipeline}
\end{figure}

\begin{figure}[h]
  \centering
\includegraphics[width=0.47\textwidth]{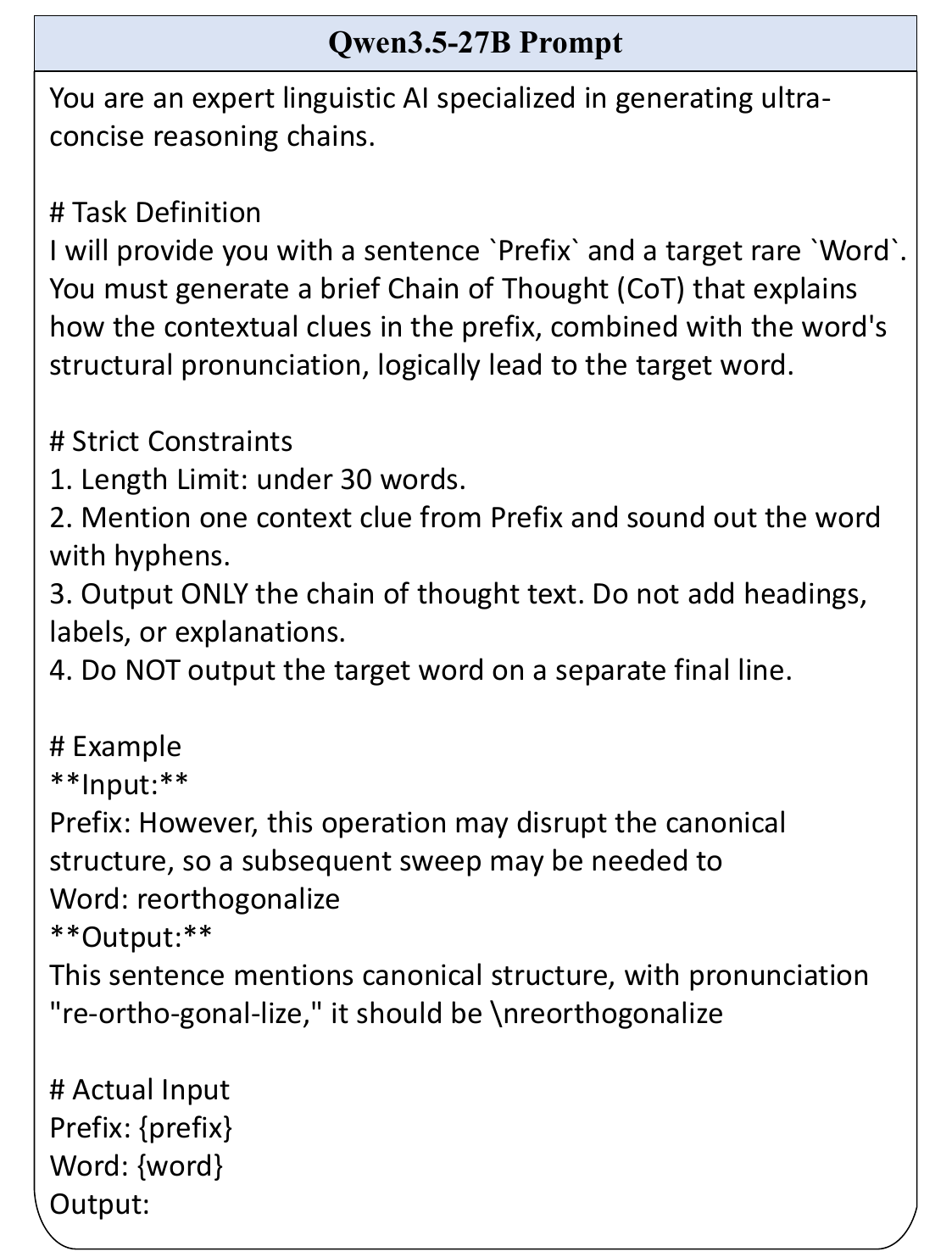}
  \caption{Qwen3.5-27B prompt for CoT generation.}
  \label{fig:cot_prompt}
\end{figure}

\noindent The construction pipeline of Spoken Darwin-Science is shown in Figure~\ref{fig:detailed_pipeline}. The rare text corpus contains extensive papers related to 440B tokens, and we extract the main text by removing the sentences at the front and end of around 200 characters each. The remaining text is further segmented into sentences using NLTK, and sentences that contain only common words are filtered, using the GigaSpeech vocabulary (79,892 words). The retained sentences are deduplicated, and speech synthesis is performed using 814 different voice clips. The generated speech is compared with the estimated duration, and excessively long audio caused by model errors is removed. After that, we select 20\% of the data for each subject, perform forced alignment to obtain the start and end timestamps of the terminology, and use Qwen3.5-27B to obtain the CoT based on the preceding history of the terminology. The prompt to generate CoT based on the previous context of the terminology is shown in Figure~\ref{fig:cot_prompt}.

\section{Spoken Darwin-Science Statistics} \label{app: dataset}

The number of instances and total duration of Spoken Darwin-Science are shown in Table~\ref{tab: app_dataset_statisitics}. The full training corpus consists of over 6,000 hours of speech and around 2.7M instances, and the 20\% subset we randomly selected is highly aligned with its distribution. The evaluation set consists of 2,000 real-world speech samples, totaling 6.081 hours.

\begin{table}[htbp]
\centering
\begin{tabular}{lccc}
\hline
\textbf{Set} & \textbf{Instance}  & \textbf{Dur (h)} \\ 
\hline
Train (full) &  2,695,953 & 6424.337  \\ 
Train (20\%) & 539,082 & 1284.519 \\ 
Evaluation &  2,000 & 6.081 \\ 
\hline
\end{tabular}
\caption{Dataset Distribution}
\label{tab: app_dataset_statisitics}
\end{table}

The subject distribution statistics of the 20\% training set are shown in Table~\ref{tab: app_subject_statisitics}. Most terminologies only appear 1-2 times.

\begin{table*}[htbp]
\centering
\begin{tabular}{lcccc}
\hline
\textbf{Subject} & \textbf{Instance Num}  & \textbf{Dur (h)} & \textbf{Terminology} & \textbf{Unique Terminology} \\
\hline
Medicine & 287,390 & 696.058 & 296,113 & 160,900 \\
Engineering & 47,400 & 112.820 & 48,794 & 38,112 \\
Biology & 45,727 & 108.823 & 47,291 & 37,762 \\
STEM (Others) & 43,953 & 102.945 & 45,706 & 36,834 \\
Human Social & 31,658 & 75.279 & 33,058 & 27,218 \\
Chemistry & 23,782 & 55.820 & 24,501 & 19,959  \\
Computer Science & 24,015 & 54.731 & 24,705 & 20,241 \\
Mathematics & 20,112 & 43.383 & 20,488 & 16,279 \\
Physics & 15,045 & 34.659  & 15,349 & 12,749 \\
\hline
\end{tabular}
\caption{Subject Distribution}
\label{tab: app_subject_statisitics}
\end{table*}

\section{Human Verification} \label{app: human_verification}

\begin{figure*}[h]
  \centering
\includegraphics[width=0.88\textwidth]{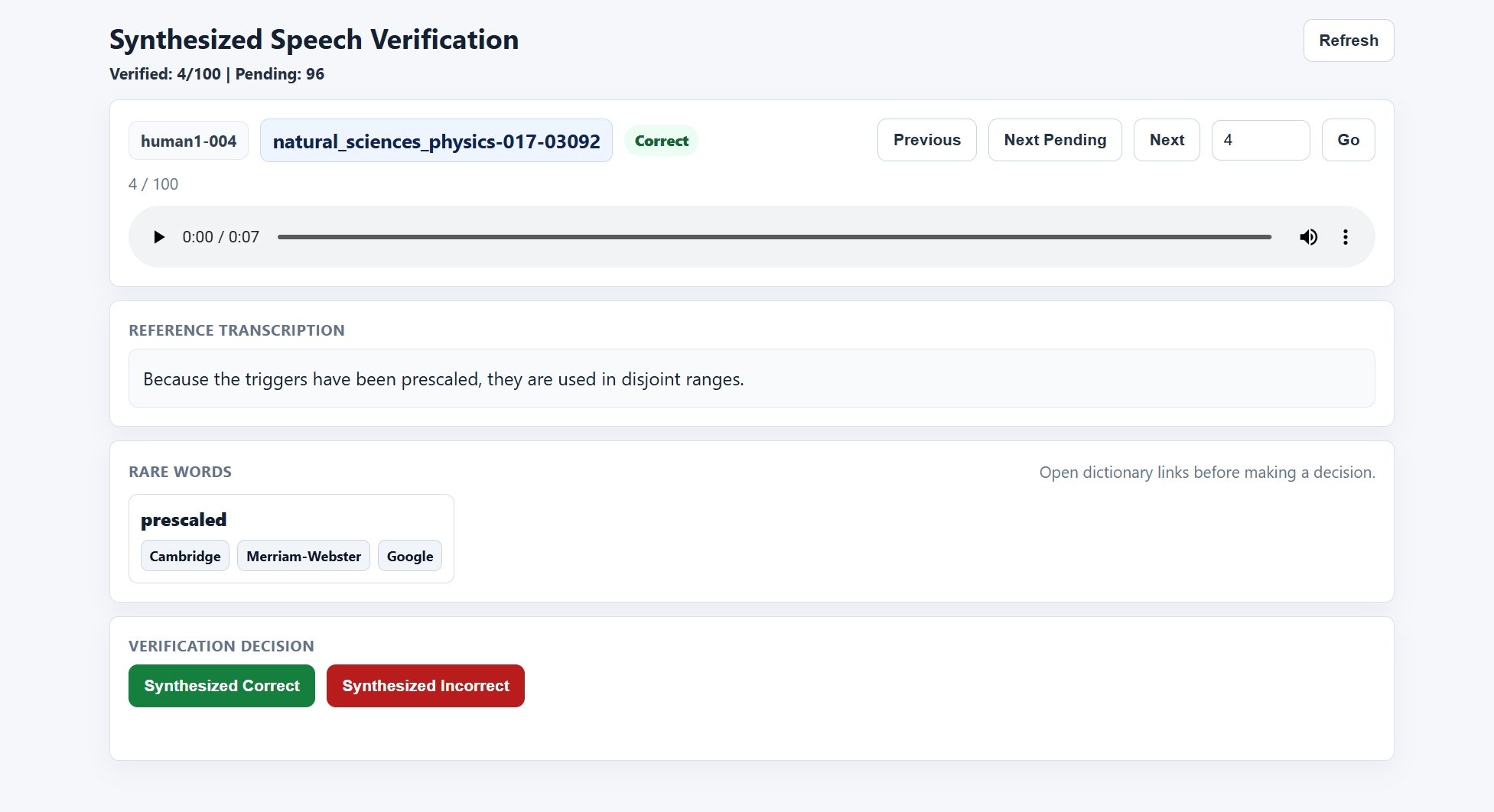}
  \caption{The screenshot of human verification.}
  \label{fig:human_verification}
\end{figure*}

Four students fluent in English were guided to participate in the manual verification work. All of them are currently pursuing undergraduate or graduate degrees. Considering the wide range of subjects covered in Spoken Darwin-Science and the difficulty of pronunciation, we provided links to the Cambridge and Merriam-Webster online dictionaries for pronunciation correction in terminology; in addition, Google search links were provided for easy access to additional information. The screenshot of the verification process is shown in Figure~\ref{fig:human_verification}.

\section{Supervised Fine-tuning Performance} \label{app:sft_exp}

\begin{figure}[t]
  \centering
\includegraphics[width=0.46\textwidth]{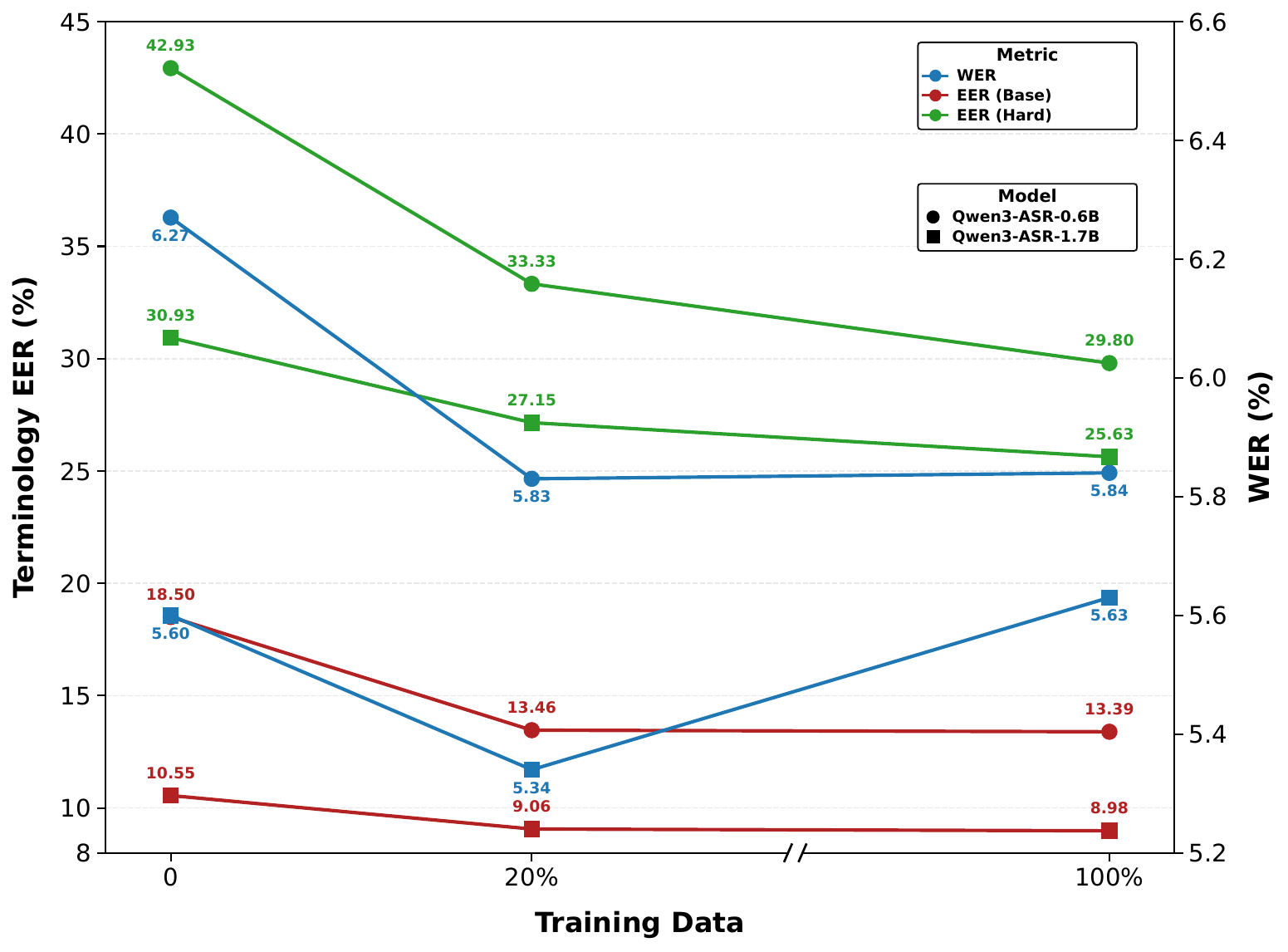}
  \caption{Supervised finetuning results on Qwen3-ASR.}
  \label{fig:sft_qwen3asr}
\end{figure}

The detailed statistics of fine-tuned LLM-based ASR models are shown in Figure~\ref{fig:sft_qwen3asr}. The terminology recognition performance of both models improves as the training data volume increases, although only marginal gains are observed when scaling from 20\% to the full training set. The overall recognition performance is stronger with only 20\% of the training data, which may be attributed to the inclusion of a large number of rare and unfamiliar terms in the full training set. These terms may alter the distribution of the training vocabulary, potentially compromising the model's recognition performance on more common terminology. Under the 20\% training data setting, the EER of the 0.6B model on the Hard set decreases from 42.93\% to 33.33\%, corresponding to a 22.36\% relative improvement, while the EER on the Base set drops from 18.50\% to 13.46\%, corresponding to a 27.24\% relative improvement. With the full training data, the EERs on the Hard set are 29.80\% and 25.63\%, respectively. The 1.7B model consistently outperforms the 0.6B model across all settings, while exhibiting a similar performance trend as the 0.6B model.

\begin{table*}[htbp]
\centering
\begin{tabular}{cccccccc}
\hline
\textbf{Top-K} & \textbf{SFT (\%)} & \textbf{LaSR (\%)} & \textbf{Absolute Gain ($\Delta$, \%)}  & \textbf{Sign-Test $p$-value} & \textbf{LaSR-wins / flips} \\
\hline
50 & 17.8 & 25.6 & +7.8 & $4.7 \times 10^{-5}$ & 68/97 \\
100 & 22.6 & 33.2 & +10.6 & $1.2 \times 10^{-6}$ & 89/125 \\
200 & 30.2 & 40.4 & +10.2 & $6.7 \times 10^{-6}$ & 93/135 \\
500 & 37.2 & 50.8 & +13.6 & $2.9 \times 10^{-9}$ & 103/138 \\
\hline
\end{tabular}
\caption{Anchor-Frame Recognition Rate at Layer-32.}
\label{tab: mechanism_tab1}
\end{table*}

\begin{table*}[htbp]
\centering
\begin{tabular}{clccccc}
\hline
\textbf{Subset} & \textbf{SFT} & \textbf{LaSR} & \textbf{Absolute Gain ($\Delta$, \%)}  & \textbf{Sign-Test $p$-value} \\
\hline
Base & 23.2\% & 32.8\% & +9.6 & $9.2 \times 10^{-4}$  \\ 
Hard & 22.0\% & 33.6\% & +11.6 & $3.2 \times 10^{-4}$ \\
\hline
\end{tabular}
\caption{Robustness across different difficulty levels.}
\label{tab: mechanism_tab2}
\end{table*}

\begin{table}[t!]
\centering
\begin{tabular}{cccc}
\hline
\textbf{subject} & \textbf{Metric} & \textbf{SFT} & \textbf{LaSR}  \\
\hline
Rank & Median & 1368 & 470 \\
Rank & Mean & 9172 & 5518 \\
Logit & Mean & 11.09 & 17.22 \\ 
\hline
\end{tabular}
\caption{Robustness across different difficulty levels.}
\label{tab: mechanism_tab1_1}
\end{table}

\section{LaSR Mechanism}\label{app:mechanism}

To thoroughly justify why LaSR works, we conducted a rigorous Logit-Lens probing experiment. We used a probe set of 500 test samples containing academic terminology, completely disjoint from the training data. The rare-word temporal anchors are derived using Qwen3-ForcedAligner-0.6B on test audio. We extracted the hidden states of both the standard SFT baseline and LaSR (-0.50s) models at the terminology anchor frame. These states are projected through the frozen text lm\_head across layers to evaluate the logit and rank shift of the target subwords. A non-parametric paired sign-test is applied to ensure statistical rigor. The probing results provide direct mechanistic evidence that LaSR fundamentally reshapes the internal representations in the deep layers (specifically emerging sharply at layers 28–32).

\subsection{Anchor-Frame Recognition Rate}

We measure the fraction of samples whose target terminology subword successfully enters the decoder's Top-k predictions at the anchor frames at Layer-32. As shown in Table~\ref{tab: mechanism_tab1}, 33.2\% of instances of the model with LaSR enter the Top-100, while this statistic for the model with SFT is only 22.6\%. Furthermore, over half of the targeted subwords ranks Top-500 with LaSR, compared to only 37.2\% after standard SFT.


For the targeted subwords, we further report the median and mean rank, as well as the average target logit in Table~\ref{tab: mechanism_tab1_1}. LaSR improves the median rank of the optimal token from 1,368 to 470, accompanied by a substantial increase in its corresponding logit. Notably, this effect is difficult to observe in the shallow layers but becomes increasingly pronounced in the deeper layers. Among the 138 samples exhibiting changes across the 500 speech inputs, 103 move in the direction favorable to LaSR. This provides a mechanistic signature of an anchor-prefix representation that is pre-conditioned to carry information about the target word before the decoder emits it.

\subsection{Difficulty Comparison}

To verify that this mechanism isn't merely inflating frequent tokens, we broke down the Layer-32 Top-100 recognition rate by terminology difficulty, as listed in Table~\ref{tab: mechanism_tab2}. Both base and hard subsets comprise 250 instances. These results indicate that the reshaping effect is at least as strong on the OOV hard subset as on the easier rare words, supporting the idea that LaSR improves contextual reasoning for low-frequency terminology rather than inflating frequent tokens.

\section{Inference Latency Analysis}\label{app: latency}

\begin{table*}[t!]
\centering
\begin{tabular}{clccccc}
\hline
\textbf{Model} & \textbf{Mode} & \textbf{RTF} & \textbf{WER (\%)}  & \textbf{EER (Base, \%)} &  \textbf{EER (Hard, \%)} &  \textbf{EER (All, \%)} \\
\hline
Base & 1-Forward & 0.1033 & 6.25 & 12.68 & 32.74 & 20.37 \\ \hline
A & 1-Forward & 0.1036 & 6.09 & 10.79 & 25.00 & 16.23 \\
A & N-Forward & 0.2705 & 6.06 & 10.63 & 24.87 & 16.08  \\ \hline
B & 1-Forward & 0.1037 & 6.17 & 11.10 & 25.13 & 16.42 \\
B & N-Forward & 0.2706 & 6.17 & 11.34 & 25.00 & 16.52 \\
\hline
\end{tabular}
\caption{Model performance under different inference modes.}
\label{tab: latency_analysis}
\end{table*}

\subsection{LaSR Inference Latency}

At inference time, LaSR does not require explicit generation of the latent chain-of-thought sequence. The latent supervision is introduced only during training by assigning CoT targets to selected audio-placeholder positions, thereby shaping the internal speech-context representations of the backbone. For non-streaming ASR decoding, the complete audio prompt can therefore be prefetched in a single causal forward pass before autoregressive transcription generation. This standard batched-prefill decoding is theoretically equivalent to sequentially consuming the prompt with a KV cache under the same causal attention pattern, but is substantially more efficient in wall-clock time. Since the sequential latent-style decoder does not explicitly decode or feed back latent CoT tokens, it is not expected to provide an intrinsic accuracy advantage over standard speech-to-text (S2T) decoding; any performance difference mainly arises from implementation details such as numerical effects, stopping criteria, and cache-based prompt processing.

The comparison of these two decoding schemes is located in Table ~\ref{tab: latency_analysis}. We select the baseline model and two random model checkpoints, corresponding to Rows 2 and 3 with LaSR of Table~\ref {tab: result1_main}, respectively. The difference in recognition accuracy between the two modes is negligible. In addition, we report the real-time factor (RTF) for both decoding schemes, defined as decoding time divided by speech duration. All experiments are performed on a single NVIDIA L20 GPU, averaging the first 100 evaluation instances after an additional 10 instances for GPU warm-up.

\begin{table}[t]
\centering
\resizebox{0.5\textwidth}{!}{
\begin{tabular}{lccccc}
\toprule
\textbf{Method} & \textbf{Token/Instance} & \textbf{RTF} & \textbf{CoT Latency (s)} \\
\midrule
CoT-ASR & 81.47 & 0.2227 & 1.304 \\
LaSR & 37.90 & 0.1036 & 0 \\
\bottomrule
\end{tabular}
}
\caption{Latency comparison with CoT-ASR.}
\label{tab:latency_compare}
\vspace{-3mm}
\end{table}

\subsection{Comparison with CoT-ASR}

CoT-ASR requires the reasoning trajectory to be explicitly generated and displayed before the model produces the transcription, making it less suitable for seamless, real-time speech recognition and interaction. We compare LaSR with CoT-ASR on the test set, as shown in Table~\ref{tab:latency_compare}. CoT-ASR first generates reasoning tokens that exceed the whole length of the transcriptions, introducing an average latency of 1.304s during inference on an L20 GPU. In contrast, LaSR does not require explicit reasoning outputs, resulting in zero CoT latency.

\section{Latent Reasoning Period Ratio}



\begin{figure}[h]
  \centering
\includegraphics[width=0.5\textwidth]{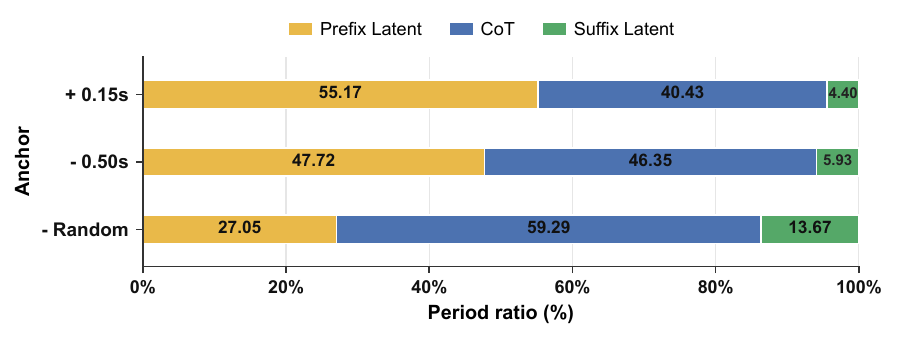}
  \caption{Average period ratio in our anchor strategies.}
  \label{fig:latent_ratio}
  \vspace{-2mm}
\end{figure}

\noindent In Figure~\ref{fig:latent_ratio}, we present the period ratio before transcription generation. In our setup, the prefix latent reasoning stage and the CoT region occupy the largest proportions. A longer prefix reasoning phase enables the model to better capture contextual information, thereby facilitating more plausible reasoning processes. In contrast, the suffix latent reasoning stage exhibited the smallest proportion and was even absent in certain samples. In the random pre-position experiments, extending the suffix latent reasoning stage led to a slight degradation in performance. This observation highlights the importance of allocating more reasoning capacity to contextual perception and acoustic information grounding, rather than to response planning strategies and transcriptional transition, which is mainly addressed in the interval between the end of CoT generation and the beginning of transcription.

\section{Parameter Constraints}

\begin{table}[h]
  \centering
  \begin{tabular}{ccccc}
    \hline
    \multirow{2}{*}{\textbf{Method}} & \multirow{2}{*}{\textbf{WER (\%)}} & \multicolumn{3}{c}{\textbf{Terminology EER (\%)}} \\ \cmidrule(r){3-5}
    & & \textbf{Base} & \textbf{Hard} & \textbf{All} \\
    \hline
   - & 5.60 & 10.55 & 30.93 & 18.37 \\
   SFT & \textbf{5.34} & \textbf{9.06} & \textbf{27.15} & \textbf{15.99} \\
   AR CoT & 6.26 & 22.70 & 38.38 & 28.73 \\
   LaSR & 5.83 & 10.79 & 31.69 & 18.80 \\
    \hline
  \end{tabular}
  \caption{\label{tab: qwen3_asr LaSR}
    Evaluation results on Qwen3-ASR-1.7B. Best performance is marked as \textbf{Bold}.
  }
  \vspace{-2mm}
\end{table}

\noindent LaSR requires initial reasoning capability of the backbone. This is reflected in the fact that for models with a small number of parameters, the addition of extra thinking content may increase interference and seriously affect the overall transcription performance, thus making LaSR less effective than non-thinking CoT. As shown in Table~\ref{tab: qwen3_asr LaSR}, Qwen3-ASR-1.7B is further employed. After training the model to autoregressively generate CoT thinking text before transcription (AR CoT), the LLM-based ASR model successfully generated reasoning paths, while the overall recognition and terminology recognition both deteriorated significantly. LaSR's effect is more moderate, but all metrics show some decline, and it performs worse than the non-thinking SFT. This indicates that with limited parameters and insufficient reasoning capabilities, LaSR could produce interference.

\section{Impact on Other Tasks}

This paper primarily focuses on terminology speech recognition, where the target words are predominantly named entities that are crucial for downstream tasks such as semantic understanding, scientific question answering, and dialogue generation. We further extend our experiments in two directions. First, we evaluate the general speech performance in general scenarios to broaden the ASR scope. Second, we extend the task to terminology perception in the scientific question answering task, investigating whether improved terminology recognition can potentially benefit downstream semantic understanding.

\subsection{General Speech Recognition}

In Section~\ref{subsec:zero_shot_exp}, we have demonstrated that the model trained on scientific domains with LaSR generalizes well to speech recognition in other specialized domains. In contrast, general scenarios dominated by daily-use words may not benefit from additional reasoning. Indeed, extensive studies have shown that on relatively simple tasks, thinking-based models can underperform non-thinking baselines, particularly in speech-related tasks~\cite{tian2025step, fan2026incentivizing}.

To further investigate this phenomenon, we conduct additional experiments on LibriSpeech, as shown in Table~\ref{tab:librispeech_exp}. Standard SFT consistently degrades the model performance across all four sets, primarily stemming from a mismatch in vocabulary distributions, as the additional training stages alter the distribution of the training data. The LaSR-based model performs worse than the SFT baseline, with the two models achieving comparable performance only on the test-clean set. Since the training strategy is the only variable between these models, these results suggest that excessive reasoning may introduce unnecessary complexity and confuse the model to some extent when processing relatively simple speech recognition tasks.

\begin{table}[t]
\centering
\resizebox{0.5\textwidth}{!}{
\begin{tabular}{lccc}
\toprule
\textbf{Set} & \textbf{Fun-Audio-Chat (\%)} & \textbf{+SFT (\%)} & \textbf{+LaSR (\%)}  \\    \hline
\textbf{Dev Clean} & 1.96 & 3.29 & 3.41 \\
\textbf{Dev Other} & 3.63 & 7.27 & 7.50 \\
\textbf{Test Clean} & 1.91 & 3.29 & 3.28 \\
\textbf{Test Other} & 4.03 & 6.83 & 7.61 \\ \hline
\textbf{Avg} & 2.87 & 5.14 & 5.42 \\
\bottomrule
\end{tabular}
}
\caption{Evaluation results on LibriSpeech.}
\label{tab:librispeech_exp}
\end{table}

\begin{table}[t]
\centering
\resizebox{0.5\textwidth}{!}{
\begin{tabular}{lccc}
\toprule
\textbf{Set} & \textbf{Fun-Audio-Chat (\%)}  & \textbf{+LaSR (\%)}  \\    \hline
\textbf{Core Entity} & 76.54 & \textbf{77.16}  \\
\textbf{Target Concept} & \textbf{66.78} & 63.84  \\
\textbf{Knowledge Anchors} & \textbf{75.57} & 64.66  \\
\bottomrule
\end{tabular}
}
\caption{Perception performance of different entity types on $S^3$-Knowledge-Terminology.}
\label{tab:s3_knowledge_exp}
\end{table}

\subsection{Terminology Perception in Scientific Question Answering}

  \begin{table*}[htbp]
  \centering
  \small
  \vspace{-10mm}
  \begin{tabular}{@{}p{0.15\textwidth} p{0.85\textwidth}@{}}
  \toprule
  \textbf{Source} & \textbf{Content} \\ \midrule
  \textbf{Transcription} & A hybrid worker species. This is what was assumed to be going on with Messor \textbf{ibericus}, but Jonathan couldn't find any evidence for that. \\ \\
  \textbf{CoT} & This is a scientific discussion regarding the classification of the hybrid worker species Messor \textbf{barbaricus}, where the speaker contrasts previous assumptions with new findings by researcher Jonathan. \\ \\
  \textbf{ASR Result} & A hybrid worker species? This is what was assumed to be going on with Messor \textbf{barbaricus}, but Jonathan couldn't find any evidence for that. \\ \hline
  \textbf{Transcription} & It has spherical cells which undergo something called \textbf{coenocytic} division. In this process, the nucleus divides repeatedly within a single cell until the cell comes apart. \\ \\
  \textbf{CoT} & This is an educational or scientific explanation describing the unique cellular behavior of spherical cells, specifically focusing on a process known as \textbf{syncytic} division where the nucleus divides repeatedly within a single cell until it ruptures. \\ \\
  \textbf{ASR Result} & It has spherical cells, which undergo something called \textbf{syncytic} division. In this process, the nucleus divides repeatedly within a single cell until the cell comes apart. \\
  \bottomrule
  \end{tabular}
  \caption{Bad cases from the recognition results of CoT-ASR. \textbf{Bold} indicates the targeted terminology.}
  \label{tab:bad_case_examples}
  \vspace{-4mm}
  \end{table*}

Since the Fun-Audio-Chat variant is fine-tuned solely in an ASR setting, directly evaluating it on scientific question answering would be unfair. We adopt the perception dimension of $S^3$-Knowledge-Terminology~\cite{liu2026s3} as a compromise. In its open-ended question answering setting, it focuses on the entities recognized in the model output and compares them against a predefined vocabulary of associated terms, including core entities, target concepts, and knowledge anchors. Core entities represent the key named entities explicitly mentioned in the question, while the latter two categories capture, to different extents, concepts associated with knowledge reasoning and key elements along the corresponding reasoning chains. As listed in Table~\ref{tab:s3_knowledge_exp}, the model trained with LaSR-based ASR achieves improved perception performance on core entities compared with the base model, while its performance on the other two categories decreases substantially. This result suggests that LaSR strengthens the capabilities most directly associated with terminology recognition, whereas the association with extrapolated concepts is weakened by the constraints of the training task. These results demonstrate the potential of LaSR to generalize beyond speech recognition to downstream tasks by adjusting training tasks and reasoning trategies.

\section{Compare with Explicit CoT (CoT-ASR)}

LaSR places CoT supervision before the model outputs. In contrast, CoT-ASR explicitly generates and displays the reasoning trajectory before producing the transcription, with the full contextual information already available when the reasoning process begins. In our main experiments, LaSR demonstrates a clear advantage in reasoning efficiency. Counterintuitively, it also achieves slightly better recognition accuracy than CoT-ASR. In this section, we compare the differences between our implementation and experiments of CoT-ASR~\cite{deng2026speech} and investigate the potential factors that may account for the performance gap.

\begin{itemize}
    \item \textbf{Training Data Scale:} The experiments in CoT-ASR use 38,000 hours of anonymized internal English ASR data, whereas our comparison uses only 1,285 hours. Training on our scope takes approximately 33 hours on 8 A100 GPUs, implying a prohibitively high computational cost if scaled to the data volume used by CoT-ASR. Moreover, speech interaction models such as Fun-Audio-Chat are generally trained without a thinking mode. Under a limited training scale, explicit CoT fine-tuning may therefore struggle to override the model's existing tendency. In contrast, LaSR incorporates reasoning into the encoding-side prior, enabling the model to benefit from reasoning without requiring an explicit reasoning trajectory during decoding, and consequently achieves comparable or even better recognition performance.

    \item \textbf{Initial Model:} CoT-ASR fine-tunes the text LLM Phi-4-mini-instruct~\cite{abouelenin2025phi}, whereas LaSR starts from a well-trained Speech LLM. Explicit CoT training has been shown to cause performance degradation in Speech LLMs~\cite{tian2025step, fan2026incentivizing}. However, since CoT-ASR starts from a text-based model and is trained exclusively on data containing reasoning chains, this issue may be alleviated.

    \item \textbf{Entity Definition:} CoT-ASR does not disclose the entity definitions used in its training and test data, but its illustrative CoT examples include the software application "BitWarden". Such proper nouns differ from the scientific content terms considered in LaSR. Entities such as "BitWarden'' generally require more comprehensive contextual information to infer their underlying concepts, whereas scientific terminology can often be identified from a small number of keywords that indicate the relevant topic. Therefore, LaSR can achieve strong performance by performing reasoning from only partial context.
    
    \item \textbf{Excessive focus on semantics:} Regarding the decoding behavior of CoT-ASR, we hypothesize that its relatively weaker performance may stem from an excessive emphasis on semantic interpretation at the expense of pronunciations. In contrast, LaSR incorporates both contextual information and pronunciation-related constraints. This observation is supported by an analysis of bad cases from the CoT-ASR model in Table~\ref{tab:bad_case_examples}, where the generated reasoning trajectories tend to prioritize semantic associations while overlooking pronunciation cues that are important for accurately recognizing the terminology.
    
\end{itemize}



\end{document}